\documentclass{article}
\usepackage{ijcai21}

\usepackage{times}
\usepackage{soul}
\usepackage{url}
\usepackage[hidelinks]{hyperref}
\usepackage[utf8]{inputenc}
\usepackage[small]{caption}
\usepackage{graphicx}
\usepackage{amsmath}
\usepackage{amsthm}
\usepackage{booktabs}
\usepackage{algorithm}

\usepackage{amssymb}
\usepackage{pifont}
\usepackage{paralist}
\usepackage{enumerate}
\usepackage{relsize}
\usepackage{xcolor}
\usepackage{subfigure}
\usepackage{caption}

\usepackage{framed,multirow}
\usepackage[noend]{algpseudocode}
\usepackage[switch]{lineno}
\usepackage{algorithmicx}  

\makeatletter
\newcommand{\printfnsymbol}[1]{%
  \textsuperscript{\@fnsymbol{#1}}%
}
\makeatother

\title{Stochastic Actor-Executor-Critic for Image-to-Image Translation}

\author{
Ziwei Luo$^1$\thanks{Equal contribution.}\and
Jing Hu$^1$\printfnsymbol{1}\and
Xin Wang$^{2}$\thanks{Corresponding authors.}\and
Siwei Lyu$^3$\and
Bin Kong$^2$\and\\
Youbing Yin$^{2}$\and
Qi Song$^{2}$\And
Xi Wu$^{1}$\printfnsymbol{2}\\
\affiliations
$^1$Chengdu University of Information Technology, China\\
$^2$Keya Medical, Seattle, USA\\
$^3$University at Buffalo, SUNY, USA\\
\emails
xi.wu@cuit.edu.cn,
xinw@keyamedna.com
}

\begin{document}
\maketitle

\begin{abstract}
  
Training a model-free deep reinforcement learning model to solve image-to-image translation is difficult since it involves high-dimensional continuous state and action spaces. In this paper, we draw inspiration from the recent success of the maximum entropy reinforcement learning framework designed for challenging continuous control problems to develop stochastic policies over high dimensional continuous spaces including image representation, generation, and control simultaneously. Central to this method is the {\em Stochastic Actor-Executor-Critic} (SAEC) which is an off-policy actor-critic model with an additional executor to generate realistic images. Specifically, the actor focuses on the high-level representation and control policy by a stochastic latent action, as well as explicitly directs the executor to generate low-level actions to manipulate the state. 
Experiments on several image-to-image translation tasks have demonstrated the effectiveness and robustness of the proposed SAEC when facing high-dimensional continuous space problems. 
\end{abstract}

\section{Introduction}

Many computer vision problems, such as face inpainting, semantic segmentation, and realistic photo generated from sketch, can be defined as learning image-to-image  translation (I2IT). Currently the most effective I2IT solution is based on a one-step framework that generates images in a single run of a deep learning (DL) model, such as VAE, U-Net, and conditional GANs. Directly learning I2IT with these DL models is challenging, due to the abundance of local minimums and poor generalization caused by overfitting. Although these problems could be  potentially alleviated by using multi-scale models or a multi-stage pipelines, we are still left with models that have intrinsically high complexities, for which the optimal parameters (e.g. stage number and scale factor) have to be determined in a subjective and {\em ad hoc} manner. 

To address these limitations of DL-based methods, we explore solving I2IT problems by leveraging the recent advances in deep reinforcement learning (DRL). The key idea is to decompose the monolithic learning process into small steps by a lighter-weight CNN, with the aim of progressively improving the quality of the model. Although recent works have successfully applied DRL to solve several visual tasks \cite{caicedo2015active}, the action space in their tasks is usually discrete and 
can not be used for I2IT that requires continuous action spaces.


Currently, a promising direction for learning continuous actions is the maximum entropy reinforcement learning (MERL), which improves both exploration and robustness by maximizing a standard RL objective with an entropy term \cite{haarnoja2018soft}. Soft actor-critic (SAC) \cite{haarnoja2018soft} is an instance of MERL and has been applied to solve continuous action tasks. 
However, the main issue hindering the practical applicability of SAC on I2IT is its inability to handle high-dimensional states and actions effectively. Although recent work \cite{yarats2019improving} addresses this problem by combining SAC with a regularized autoencoder (RAE), RAE only provides an auxiliary loss for an end-to-end RL training and is incapable for I2IT tasks. 

Besides, a glaring issue is that
the reward signal is so sparse that leads to an unstable training, and higher dimensional
states such as pixels worsen this problem \cite{yarats2019improving}. Thus, training an image-based RL model requires much more exploration and exploitation. One solution to stablize training is to extract a lower dimensional visual representation with a separately pre-trained DNN model, and learn the
value function and corresponding
policy in the latent spaces \cite{nair2018visual}. 
However, this approach can not be trained from scratch, which could lead to inconsistent state representations with an optimal policy. Previous works like \cite{lee2020slac} have shown that it is beneficial to learn the image representations and continuous controls simultaneously.

In this paper, we propose a new DRL architecture, {\em stochastic actor-executor critic} (SAEC), to handle I2IT with very high dimensional continuous state and action spaces. SAEC is formed by three core deep neural networks: the actor, the critic, and the executor (see Fig. \ref{dl-rl}). 
High level actions are generated by the actor in a low dimensional latent space, and forwarded to the executor for image translation. The critic evaluates the latent actions to guide the policy update. These three neural networks form two different branches in SAEC. More specifically, 
the actor and the critic form the RL branch as an actor-critic model, while The actor and executor form the DL branch similar to autoencoder, with an important difference that skip connections are added between the actor and executor.
The RL branch is trained based on the MERL framework, while the DL branch is trained by minimizing a reconstruction objective. 

Our contributions can be summarized as follows:

\begin{itemize}

\item A new DRL framework {\em stochastic actor-executor critic} (SAEC) is proposed to handle very high dimensional state and action spaces in I2IT task.

\item Even using a sparse reward signal, our SAEC model can be stably trained from scratch on I2IT problems by combining DL-based supervision.

\item  Our SAEC framework is flexible to incorporate many advanced DL methods for various complex I2IT applications. 
And this framework enables the agent to simultaneously learn feature representation, control, and image generation in high-dimensional continuous spaces. 
\end{itemize}



\section{Background}


\subsection{Image-to-Image Translation}
\begin{figure*}[t]
    \centering
    \includegraphics[scale=0.545]{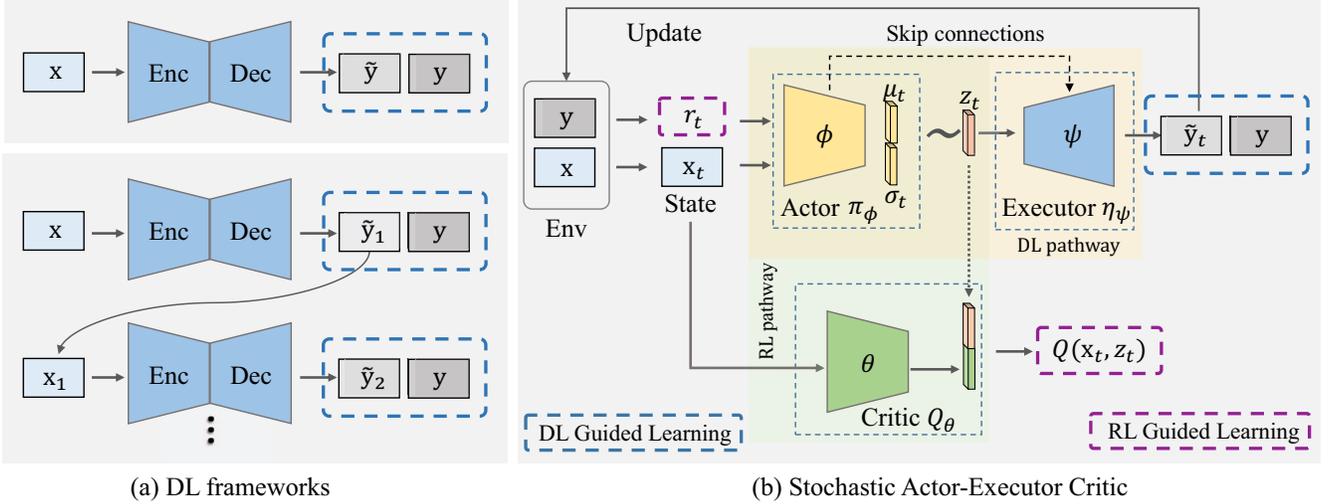}
    \caption{The conventional DL frameworks and our SAEC framework. Top of (a): One-stage DL framework. Bottom of (a): Multi-stage DL framework, whose networks are usually different in each stage. (b) Our SAEC framework. $\bf{x}$ and $\bf{y}$ are input and target, $\tilde{\bf y}$ is the predicted output. $\bf{x_t}$ and $\tilde{\bf y}_t$ are temporary inputs and outputs generated in step (stage) $t$. In the SAEC framework, there are two actions: high-level latent-action ${\bf z}_t$ which is sampled from policy $\pi_\phi$: ${\bf z_t} \sim \pi_\phi({\bf z}_t|{\bf x}_t)$, and low-level image-action $\tilde{\bf y}_t$ which is generated from executor $\eta_\psi$ conditioned by ${\bf z}_t$: $\tilde{\bf y}_t=\eta_{\psi}(\tilde{\bf y}_t|{\bf x_t}, {\bf z_t})$. Both actions are high-dimensional continuous. 
    }
    \label{dl-rl}
\end{figure*}

Image-to-image translation (I2IT) is often addressed by learning a generative process $G$ that maps state $\bf x$ to target $\bf y$, $G:{\bf x} \rightarrow {\bf y}$. The state should be consistent with the target and both of them are images,  such as generating realistic photos from semantic segmentation labels \cite{isola2017image}, or synthesizing completed visually targets from images with missing regions \cite{pathak2016context}. Autoencoder is leveraged in most research work to learn this process by minimizing the reconstruction error ${\cal L}_{rec}$ between the predicted image $\tilde{\bf y}$ and the target $\bf y$ (see Figure \ref{dl-rl}). In addition, the generative adversarial network (GAN) is also vigorously studied in I2IT to synthesis realistic images \cite{isola2017image}.
Subsequent works enhance I2IT performance by using a coarse-to-fine deep learning framework \cite{yu2018generative} that recursively uses the previous {stage's} output as the input to the next stage (see the bottom of the Fig. \ref{dl-rl}(a)). In this way, I2IT task is transformed into a multi-stage, coarse-to-fine solution. Although iteration can be infinitely applied in theory, it is limited by the increasing model size and training instability.

\subsection{Reinforcement Learning}
Reinforcement learning (RL) is an infinite-horizon Markov decision process (MDP), defined by the tuple $({\cal X},{\cal A}, {P},{r}, {\gamma})$. ${\cal X}$ is a set of states, ${\cal A}$ is action space, and $P$ represents the state transition probability given $x \in {\cal X}$ and ${\bf a}$, $r$ is the reward emitted from each transition, and $\gamma \in [0,1]$ is the reward discount factor. The standard objective of RL is to maximize the expected sum of rewards. Maximum entropy RL (MERL) further adds an entropy term:
$ \sum\nolimits_{t = 0}^T  {{\mathbb{E}_{({\bf x}_t, {\bf a}_t)\sim {\rho}_{\pi} }}[{r_{t} + \alpha {\cal H}(\cdot | {\bf x}_t)}]} $, where $\alpha$ is a temperature parameter, ${\rho}_{\pi}$ denotes the marginal trajectory distribution induced by policy $\pi$. MERL model and has proven stable and powerful in low dimensional continuous action tasks, such as games and robotic controls \cite{haarnoja2018soft}.
However, when facing complex visual problems such as I2IT, where observations and actions are high-dimensional, it remains a challenge for RL models \cite{lee2020slac}.

\section{Method}


This work reformulates I2IT as a decision making problem. Instead of directly mapping the input to the target image in a single step, we propose to use a light-weight RL model that performs the translation progressively, where new details can be added as the translation progresses. The proposed model {\em stochastic actor-executor-critic} (SAEC) consists of three components: an actor ($\pi_{\phi}$), an executor ($\eta_{\psi}$), and a critic ($Q_{\theta}$). Specifically, the actor $\pi_{\phi}$ and the executor $\eta_{\psi}$ form a DL pathway that directs the agent learning with a DL-based objective. The same actor and the critic $Q_{\theta}$ form a DRL pathway that works as an actor-critic model. 
In SAEC, the actor generates a latent action according to a stochastic policy, so as to capture the required image translation control. The critic evaluates the stochastic policy. The executor leverages the latent action and image details to perform image translation and the generated image is then applied to the environment as the new input image at next time. 

\subsection{I2IT as MDP}
\label{sec:i2i-mdp}
The key of our method is to formulate I2IT as a Markov decision process (MDP).  
As for state ${\bf x}_t \in {\cal X}$, it is defined task specific. 
For realistic photo generation or segmentation, the state ${\bf x}_t$ is the pair of the fixed source image $I_f$ and the moving image $I_{m_t}$: ${\bf x}_t=(I_f, I_{m_t})$. A new state is generated by warping the moving image with predicted image action $\tilde{\bf y}_t$: ${\bf x}_{t+1}=(I_f, \tilde{\bf y}_t \circ I_{m_t})$. For face inpainting, the state is composed of the original image ${\bf x}$ and the synthesized missing part $\tilde{\bf y}_t$. The next state is obtained by adding the new predicted image to the missing area: ${\bf x}_{t+1}=(\tilde{\bf y}_{t} \odot {\bf x}_{t})$.

There are two types of actions in our action space: latent action ${\bf z}_t \in {\cal Z}$ and image action $\tilde{\bf y}_t \in {\cal Y}$. These two actions are intrinsically different, in that $\cal Z$ has a high-level of abstraction and its distribution is unknown. ${\bf z}_t$ is sampled from the stochastic policy of the actor $\pi_{\phi}$, ${\bf z}_t \sim \pi_{\phi}({\bf x}_t)$, while $\tilde{\bf y}_t$ is created by the executor based on  ${\bf z}_t$ and the input image. The reward function is flexible and can be any evaluation metric that measures the similarity between the predicted image $\tilde{\bf y}_t$ and the target ${\bf y}$, such as the peak signal-to-noise ratio (PSNR) and the structural similarity index measure (SSIM).

The sequence of actions are implemented by the {\em stochastic actor-executor-critic} (SAEC) model. Specifically, at each time step $t$, the actor samples a latent action ${\bf z}_t$ from state ${\bf x}_t$, then the executor generates image action $\tilde{\bf y}_t$ from the latent action ${\bf z}_t$ and state ${\bf x}_t$. The change of the state will result in a reward $r_t$, and the new state is fed back to the SAEC model to generate a new image action. 
Let ${\cal T}_t$ represents a single run from state ${\bf x}_t$ to image action $\tilde{\bf y}_t$ via an actor and an executor: 
${\bf z}_t \sim \pi_{\phi}({\bf x}_t), \tilde{\bf y}_t = \eta_{\psi}({\bf x}_t, {\bf z}_t)$.
The whole process of I2IT in MDP can be formulated as follows:

\begin{equation}
\centering
\label{multi-steps}
    {\cal T}({\bf x}) = {\cal T}_t \circ {\cal T}_{t-1} \circ \dotsm \circ {\cal T}_0({\bf x}),
\end{equation}
where $\circ$ is the function composition operation, ${\cal T}_t$ is the $t$th translation step that predicts image from current state ${\bf x}_t$.

\subsection{Policy with Skip Connections}
\label{sec:su-net}

As in an MERL model, the stochastic policy and maximum entropy in our SAEC improve the exploration for more diverse generation possibilities, which helps to prevent agent from producing a single type of plausible output during training (known as mode-collapse). In addition, one specific characteristic in the SAEC model is that skip connections are added from each down-sampling layer of the actor to the corresponding up-sampling layer of the executor, as shown in Fig. \ref{dl-rl}(b). In this way, a natural looking image is more likely to be reconstructed since the details of state ${\bf x}_t$ can be passed to the executor by skip connections. Besides, since both ${\bf z}_t$ and ${\bf x}_t$ can be used by the executor to generate image action, over-exploration of the action space can be avoided in SAEC, where the variance is limited by these passed detail information. 
Furthermore, the skip connections also facilitate back-propagation of the DL gradients to the actor. Our experiments show that if the skip connections are removed, the details of the reconstructed image may lost because of down-sampling (max-pooling) and stochastic policies, and the training would be very unstable in such a situation. Combining skip connections and DL-based learning can avoid detail loss and thus stabilize training.  (See experiments section for more details). 

\subsection{Stochastic Actor-Executor-Critic}
\label{sec:saec}

To address I2IT, We propose to extend the maximum entropy RL (MERL) framework with an additional executor. Our proposed model has a similar structure with SAC \cite{haarnoja2018soft}, but the latter cannot be applied to the I2IT tasks directly since the delayed and sparse rewards making it difficult to handle the inherent high-dimensional states and actions. To this end, we propose to train the actor and executor in a conventional DL-based manner, 
where the training data are collected from the agent-environmental interaction and change dynamically in experience replay. The objective of SAEC is to learn a policy $\pi_{\phi}$ and an executor $\eta_{\psi}$, by maximizing both the conditional likelihood and the expected sum of rewards with an entropy ${\cal H}$.


\begin{algorithm}[t]
    \caption{Stochastic Actor-Executor Critic}\label{alg:sac}
    \begin{algorithmic}
        \Require Environment $ Env $ and initial parameters $\theta_1, \theta_2 $ for the Critics, $\phi, \psi$ for the Actor and Executor.
            \State $\bar{\theta}_1 \gets \theta_1, \bar{\theta}_2 \gets \theta_2, \cal D \gets \varnothing$ 
            
            \For {each iteration}
                \State ${\bf x}, {\bf y} \sim Env_{reset}()$ 
                \For {each environment step}
                    \State ${\bf z}_t \sim \pi_{\phi}({\bf z}_t|{\bf x}_t)$ 
                    \State $\tilde{\bf y}_t \gets \eta_{\psi}({\bf x}_t, {\bf z}_t)$ 
                    \State ${\bf x}_{t+1}, r_t \sim Env_{step}(\tilde{\bf y}_t)$ 
                    \State $\cal D \gets {\cal D}$ $\cup$ $\{({\bf y}, {\bf x}_t, {\bf z}_t, r_t, {\bf x}_{t+1})\}$ 
                \EndFor
                
                \For {each gradient step}
                    \State Update Actor and Executor (DL guided):
                     \State \qquad$\phi \gets \phi - \lambda_{DL} \hat{\nabla}_{\phi} {\cal L}_{DL}(\phi)$ 
                     \State \qquad$\psi \gets \psi - \lambda_{DL} \hat{\nabla}_{\psi} {\cal L}_{DL}(\psi)$
                    \State Update Actor and Critic (RL guided):

                     \State \qquad$\theta_i \gets \theta_i - \lambda_Q \hat{\nabla}_{\theta_i} J_Q(\theta_i)$ for $i \in \{1,2\}$ 
                     
                     \State \qquad$\phi \gets \phi - \lambda_{\pi} \hat{\nabla}_{\phi} (J_{\pi}(\phi))$ 
                     \State \qquad$\alpha \gets \alpha - \lambda \hat{\nabla}_{\alpha} J(\alpha)$ 
                     \State \qquad$\bar{\theta}_i \leftarrow \tau \theta_i + (1-\tau)\bar{\theta}_i$ for $i \in \{1,2\}$ 
                \EndFor
                
            \EndFor
        \Ensure $\theta_1, \theta_2, \phi, \psi $ 
        
    \end{algorithmic}
\end{algorithm}

\subsubsection{Conditional Generative Model Learning} 
The actor and the executor form a conditional generative process that translates the current state ${\bf x}_t$ to target ${\bf y}$. Instead of learning the generative model alone like a VAE, we maximize the conditional likelihood following the stochastic policy $\pi_{\phi}$:

\begin{equation}
\centering
\label{likelihood}
    p({\bf y}|{\bf x}_t)=\int \pi_{\phi}({\bf z}_t|{\bf x}_t)\eta_{\psi}({\bf y}|{\bf x}_t, {\bf z}_t)d{\bf x}_t.
\end{equation}
The empirical objective is to minimize the reconstruction error over all samples:

\begin{equation}
\centering
\label{rec_loss}
    {\cal L}_{rec}=\mathbb{E}_{{\bf x}_t,{\bf y} \sim {\cal D}}[\left \| {\cal T}({\bf x}_t),{\bf y}\right \|_d],
\end{equation}
where $\cal D$ is a replay pool, $\left \| \cdot \right \|_d$ denotes some distance measures, such as $L_1$ and $L_2$. 
To synthesis more realistic images, we also extend the actor-executor to a high quality generative model by jointly training an adversarial loss with discriminator $D$:

\begin{equation}
\centering
\label{adv_loss}
    {\cal L}_{adv}= \mathbb{E}_{{\bf x}_t,{\bf y} \sim {\cal D}}[\log (D({\bf y})) + \log (1-D({\cal T}({\bf x}_t)))]
\end{equation}
Our final DL guided objective can be expressed as
\begin{equation}
\centering
\label{dl-loss}
    {\cal L}_{DL} = {\lambda}_{rec} {\cal L}_{rec} + {\lambda}_{adv} {\cal L}_{adv},
\end{equation}
where ${\lambda}_{rec}$ and ${\lambda}_{adv}$ are used to balance the reconstruction and adversarial learning. In general, the DL pathway is very flexible and can readily leverage any other advanced DL losses or techniques into this framework. 

\subsubsection{MERL in I2IT}
In RL pathway, the rewards and the soft Q values are used to iteratively guide the stochastic policy improve. Moreover, the latent-action ${\bf z}_t$, 
is used to estimate soft state-action value for encouraging high-level policies. Let actor and critic with parameters $\phi$ and $\theta$ be the function approximators for the policy $\pi_\phi$ and the soft Q-function $Q_\theta$, respectively. The critic parameters $\theta$ are trained to minimize the soft Bellman residual: 
\begin{equation}
\centering
\label{sac-q}
\begin{aligned}
    {J_Q}({\theta}) & = \mathbb{E}_{({\bf x}_t, {\bf z}_t) \sim {\cal D}} \left[\frac{1}{2} (Q_{\theta}({\bf x}_t, {\bf z}_t) - {\hat Q}_{\bar{\theta}}({\bf x}_{t}, {\bf z}_{t}))^2\right],     
\end{aligned}
\end{equation}

\begin{equation}
    \begin{aligned}
        {\hat Q}_{\bar{\theta}}({\bf x}_{t}, {\bf z}_{t}) &= r_t + \gamma \mathbb{E}_{{\bf x}_{t+1} \sim {P}}\left[V_{\bar{\theta}}({\bf x}_{t+1})\right], \notag
    \end{aligned}
\end{equation}
where $V_{\bar{\theta}}({\bf x}_{t})$ is the soft state value function which can be computed by 

\begin{equation}
    \begin{aligned}
        V_{\bar{\theta}}({\bf x}_{t})  = \mathbb{E}_{{\bf z}_{t} \sim \pi_\phi}\left[Q_{\bar{\theta}}({\bf x}_{t}, {\bf z}_{t}) - \alpha \log \pi_\phi ({\bf z}_{t}|{\bf x}_{t})\right].
    \end{aligned}
\end{equation}
We use a target network $Q_{\bar{\theta}}$ to stabilize training, whose parameters $\bar{\theta}$ are obtained by an exponentially moving average of parameters of the critic network: $\bar{\theta} \leftarrow \tau \theta + (1-\tau)\bar{\theta}$. 
The actor parameters $\phi$ are optimized to learn the policy towards the exponential of the soft Q-function: 

\begin{equation}
\centering
\label{sac-pi}
    {J_\pi}({\phi}) = \mathbb{E}_{{\bf x}_t \sim {\cal D}} \left[\mathbb{E}_{{\bf z}_t \sim \pi_\phi}\left[\alpha \log (\pi_\phi({\bf z}_t|{\bf x}_t)) - Q_\theta({\bf x}_t, {\bf z}_t)\right]\right]
\end{equation}
In practice, we use two critics with the same structure but different parameters, ($\theta_1$, $\theta_2$), to mitigate positive bias in policy improvement and accelerate training. Moreover, we automatically tune the temperature hyperparameter $\alpha$ by 
\begin{equation}
    \begin{aligned}
        {J(\alpha)} = \mathbb{E}_{{\bf z}_t \sim \pi_t}[-\alpha \log \pi_t({\bf z}_t|{\bf x}_t) - \alpha \bar{{\cal H}}],
    \end{aligned}
\end{equation}
where $\bar{{\cal H}}$ represents a constant entropy that equals to the negative of the dimension of the latent action spaces.

The complete algorithm of SAEC is described in Algorithm \ref{alg:sac}. The training process alternates between learning I2IT from DL pathway and policy control from  RL pathway. The joint training of DL and RL ensures a fast convergence since the supervised or unsupervised loss is a "highway" to facilitate back-propagation of the gradients to the actor and the executor. As the latent-action ${\bf z}_t$ is concatenated into critic (shown in Figure \ref{dl-rl}(b)), the RL objective and DL objective can cooperate to train the stochastic actor fast and stable. 

\section{Experiments}

In this section, we demonstrate experimentally the effectiveness and robustness of the SAEC framework on several different tasks of I2IT.

\subsection{Face Inpainting}

 \noindent\textbf{Settings} 
In this experiment, we apply SAEC to the problem of face inpainting, which is to fill in pixels in the central area of a face image with synthesized contents that are semantically consistent with the original face and at the same time visually realistic. Celeba-HQ dataset is used in this study, of which $28,000$ images are used for training and $2,000$ images are used for testing. All images have a missing part of size of 64 $\times$ 64 cropped in the center. 
We compare the SAEC 
with several recent face inpainting methods, including CE \cite{pathak2016context}, CA \cite{yu2018generative}, PEN \cite{zeng2019learning}, PIC \cite{zheng2019pluralistic} and RN \cite{yu2020region}. 

The actor-executor for our methods uses the same network structure as the encoder-decoder for CE. The network structures of the discriminator also comes from CE but with a different SNGAN loss. Following the previous work \cite{pathak2016context,yu2018generative,zheng2019pluralistic}, we use PSNR and SSIM as the evaluation metrics.

\begin{figure*}[t]
    \centering
    \includegraphics[scale=0.54]{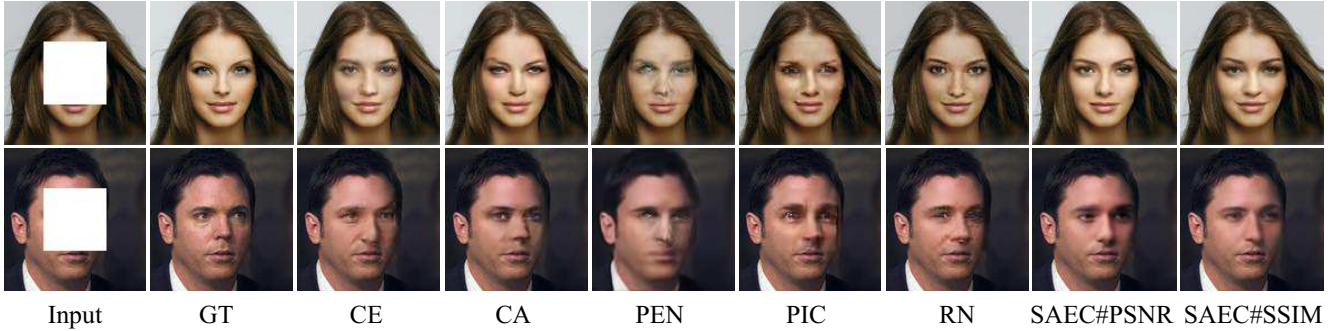}
    \caption{Visual comparison of face inpainting over all methods. Our SAEC use SNGAN for auxiliary DL guided learning. $\#$ indicates what reward is used for RL training. Our results have better visual quality even for the large pose face.}
    \label{fi-compare}
\end{figure*}

 \noindent \textbf{Results and Analysis}
As shown in Table \ref{tab:psnr}, by using $L_1$ and adversarial loss in the DL guided learning, our method achieves the best PSNR and SSIM scores comparing with the existing state-of-the-art methods. And as mentioned in the Method section, the reward function is very flexible for our RL framework. Both the PSNR and SSIM based reward are suitable for SAEC and can improve the performance on face inpainting. 
The qualitative comparison shown in Figure \ref{fi-compare} illustrates that the SAEC gives obvious visual improvement for synthesizing realistic faces. 
The results of SAEC are very reasonable, and the generated faces are sharper and more natural. This may attribute to the high-level action ${\bf z}_t$, which focuses on learning the global semantic structure and then directs the executor with DL guided learning to further improve the local details of the generated image. We can also see that the synthesised images of the SAEC could have very different appearances from the ground truth, which indicates that although our training is based on paired images, the SAEC can successfully explore and exploit data for producing diverse results.

\begin{table}
\centering
\begin{tabular}{lcc}
\toprule
\multicolumn{1}{l}{Method} & PSNR $\uparrow$ & SSIM $\uparrow$ \\ \midrule
CE \cite{pathak2016context}                         & 26.949 & 0.870     \\
CA \cite{yu2018generative}                  & 26.608  & 0.871    \\ 

PIC \cite{zheng2019pluralistic}                      & 26.307    & 0.893  \\
PEN \cite{zeng2019learning}                      & 26.391    & 0.883  \\ 
RN \cite{yu2020region}                      & 26.932    & 0.892  \\ 
\midrule
SAEC (SNGAN + PSNR reward)  & 27.176   & 0.882 \\
\textbf{SAEC (SNGAN + SSIM reward)}           & \textbf{27.327}    & \textbf{0.896} \\
\bottomrule
\end{tabular}
\captionof{table}{Quantitative results of all methods on Celeba-HQ.}
\label{tab:psnr}
\end{table}

\noindent \textbf{Ablation Study}
To illustrate the stability of training GANs in SAEC, we jointly use $L_1$ and several advanced GAN techniques i.e. WGAN-GP \cite{gulrajani2017improved}, RaGAN \cite{jolicoeur2018relativistic},
and SNGAN \cite{miyato2018spectral} for DL guided learning. To validate the effectiveness of RL pathway in SAEC for I2IT, we also separately train an actor-executor model (AE) by jointly optimizing the $L_1$ and SNGAN loss. The results shown in Table \ref{tab:fi-gans} indicate that the proposed SAEC with different GANs are stable, and significantly improve the performance of training AE with SNGAN alone, which further demonstrates the contribution of the RL formulation.

\subsection{Realistic Photo Translation}

\noindent \textbf{Settings}
In this section, we evaluate the SAEC on a set of general I2ITs.
We select three datasets of realistic photo translation including:

\begin{compactitem}

\item CMP Facades dataset for segmentation {\em labels$\rightarrow$images} \cite{tylevcek2013spatial}.

\item Cityscapes dataset for segmentation {\em labels$\rightarrow$images} and {\em images$\rightarrow$labels} \cite{Cordts2016Cityscapes}.

\item Edge and shoes dataset for {\em edges$\rightarrow$shoes} \cite{yu2014fine}.

\end{compactitem}
In these tasks, SAEC uses the same network structure as in face inpainting experiment with PSNR reward and SNGAN loss. We compare our method with pix2pix \cite{isola2017image} and PAN \cite{wang2018perceptual}.
We also compared with pix2pixHD \cite{wang2018high}, DRPAN \cite{wang2019drpan} and CHAN \cite{gao2021complementary}, which are designed for high-quality I2IT. Moreover, we replace MERL with PPO \cite{schulman2017proximal} in the RL pathway as SAEC$^*$. We use PSNR, SSIM and LPIPS \cite{zhang2018unreasonable} as the evaluation metrics.

\begin{table}[t]
\centering
\begin{tabular}{lcc}
\toprule
\multicolumn{1}{l}{Method} & PSNR $\uparrow$  & SSIM $\uparrow$  \\ \midrule
AE + SNGAN         & 26.884   & 0.871  \\ 
SAEC (+ WGAN-GP)        &27.091   & 0.875 \\
SAEC (+ RaGAN)        & 27.080    & 0.873\\
\textbf{SAEC (+ SNGAN)}                        & \textbf{27.176}   & \textbf{0.882}   \\
\bottomrule
\end{tabular}
\captionof{table}{Quantitative results of the variants methods on Celeba-HQ testing dataset (all trained with PSNR reward).}
\label{tab:fi-gans}
\end{table}

\begin{figure*}[t]
    \centering
    \includegraphics[scale=0.54]{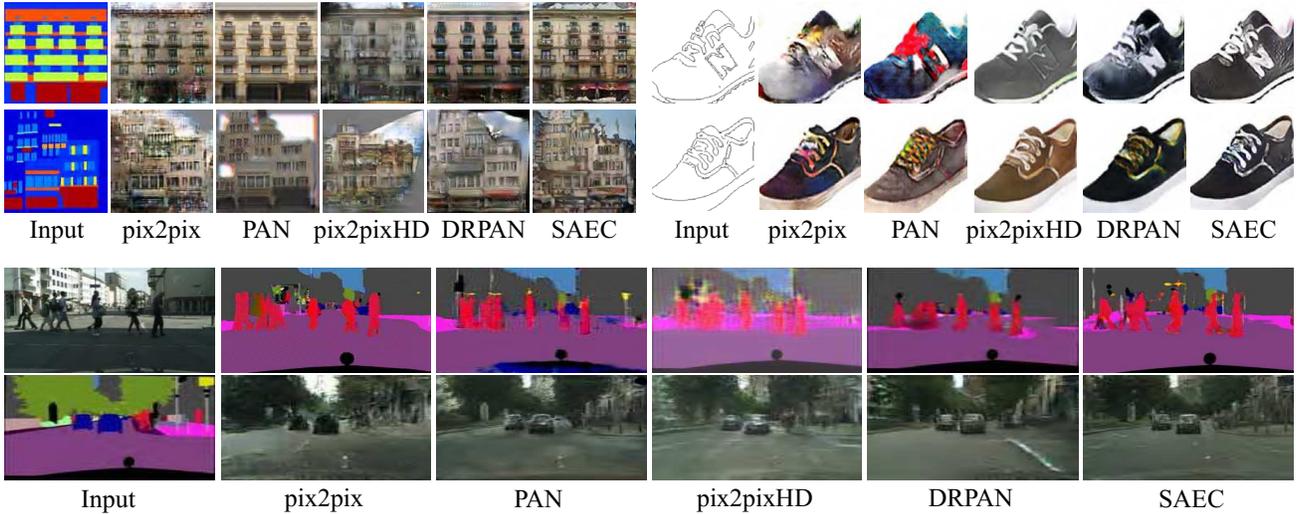}
    \caption{Visual comparison of our method with pix2pix, PAN, pix2pixHD and DRPAN over all tasks.}
    \label{l2i}
\end{figure*}

\begin{table*}[t]
\renewcommand\arraystretch{1.1}
\centering
\scalebox{0.90}{
\begin{tabular}{lcccccccccccc}

\toprule
\multirow{3}{*}{Method} & \multicolumn{3}{c}{Facades label$\rightarrow$image}   &  \multicolumn{3}{c}{Cityscapes image$\rightarrow$label}   &
\multicolumn{3}{c}{Cityscapes label$\rightarrow$image}   &\multicolumn{3}{c}{Edges$\rightarrow$shoes} \\ 
 & PSNR  & SSIM  & LPIPS $\downarrow$  & PSNR & SSIM & LPIPS $\downarrow$ & PSNR  & SSIM & LPIPS $\downarrow$ & PSNR & SSIM & LPIPS $\downarrow$ \\ \midrule
 
pix2pix         & 12.290   & 0.225 & 0.438 & 15.891   & 0.457 & 0.287  &15.193  & 0.279 & 0.379 & 15.812   & 0.625 & 0.279  \\

PAN   & 12.779     & 0.249  & 0.387  & 16.317     & 0.566    & 0.228 & 16.408   & 0.391  &0.346  &16.097 &0.658 & 0.228\\

pix2pixHD   & 12.357     & 0.162  & 0.336  & 17.606     & 0.581    & 0.204 & 15.619   & 0.361  & \textbf{0.319}  &17.110 &0.686 & 0.220\\

DRPAN   & 13.101     & 0.276  & 0.354  & 17.724     & 0.633    & 0.214 & 16.673   & 0.403  &0.343  &17.524 &\textbf{0.713} & 0.221\\

CHAN   & 13.137    & 0.231   & 0.402  & 17.459    & 0.641   & 0.222 & 16.739    & 0.401   & 0.373  & 18.065    & 0.692   & 0.236\\

\midrule
SAEC$^*$   & 13.163     & \textbf{0.308}  & 0.366  & 17.168     & 0.616    & 0.221 & 16.685   & 0.410  &0.362  &16.914 &0.695 & 0.225\\
\textbf{SAEC}   & \textbf{13.178}   & 0.296  & \textbf{0.324} & \textbf{17.969}   & \textbf{0.659}  & \textbf{0.203}  & \textbf{16.848}  & \textbf{0.412}  & 0.337  & \textbf{18.178}   & 0.698 & \textbf{0.215} \\
\bottomrule
\end{tabular}}
\captionof{table}{Quantitative results of our SAEC with other methods over all datasets. $\downarrow$ means lower is better, SAEC$^*$ means using PPO.}
\label{tab:pix2pix}
\end{table*}

\noindent \textbf{Quantitative Evaluation}
The quantitative results are shown in Table \ref{tab:pix2pix}. With a similar network structure, the proposed method significantly outperforms the pix2pix and PAN model on PSNR, SSIM and LPIPS over all datasets and tasks. The SAEC even achieves a comparable or better performance than the high-quality pix2pixHD and DRPAN model, which have much more complex architectures and training strategies. Moreover, using MERL instead of PPO obviously improves performance on most tasks. These experiments illustrate that the proposed SAEC is a robust and effective solution for I2IT.


\begin{figure*}[!h]
\centering

\subfigure{
\includegraphics[width=0.24\textwidth]{{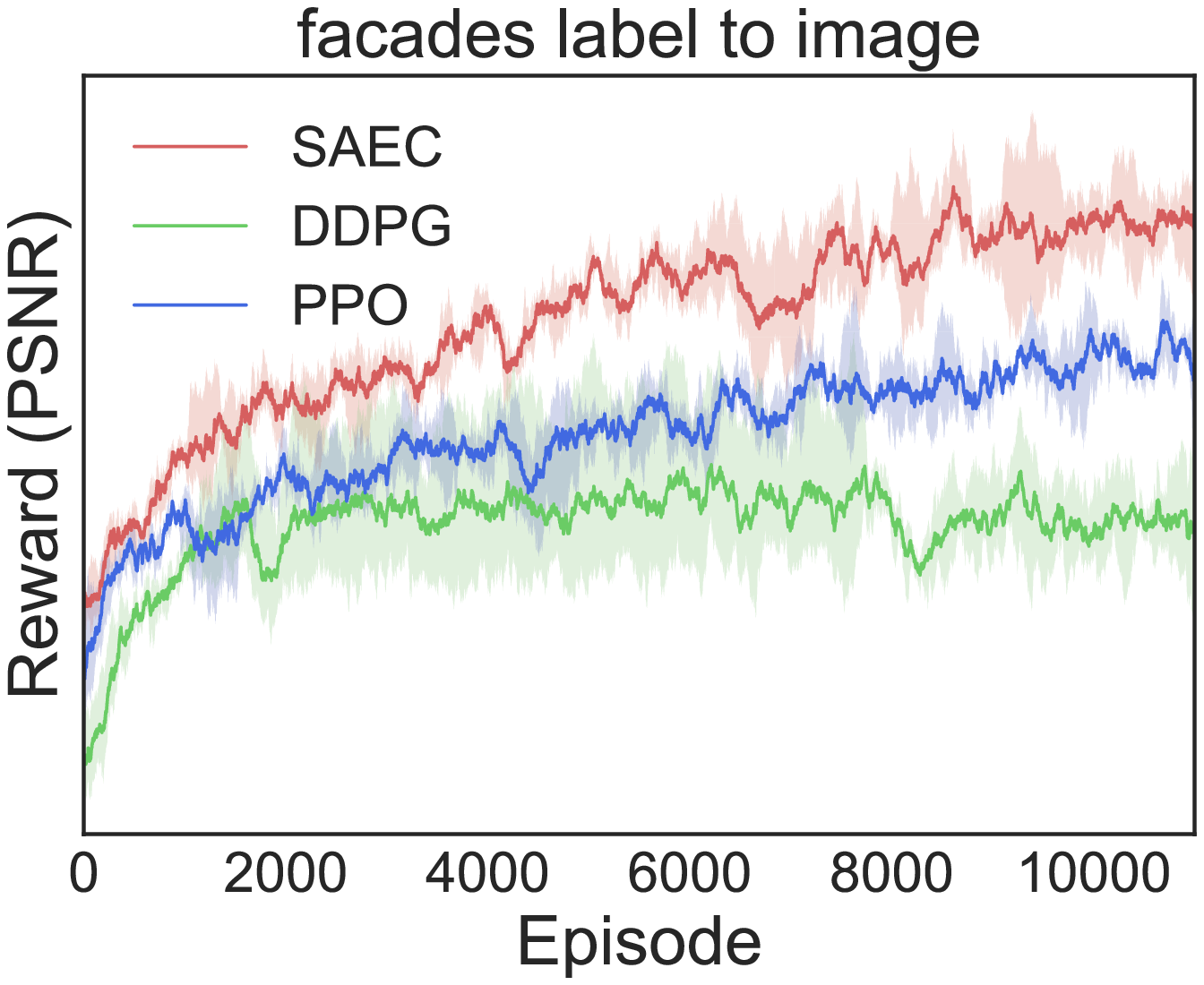}}
\includegraphics[width=0.24\textwidth]{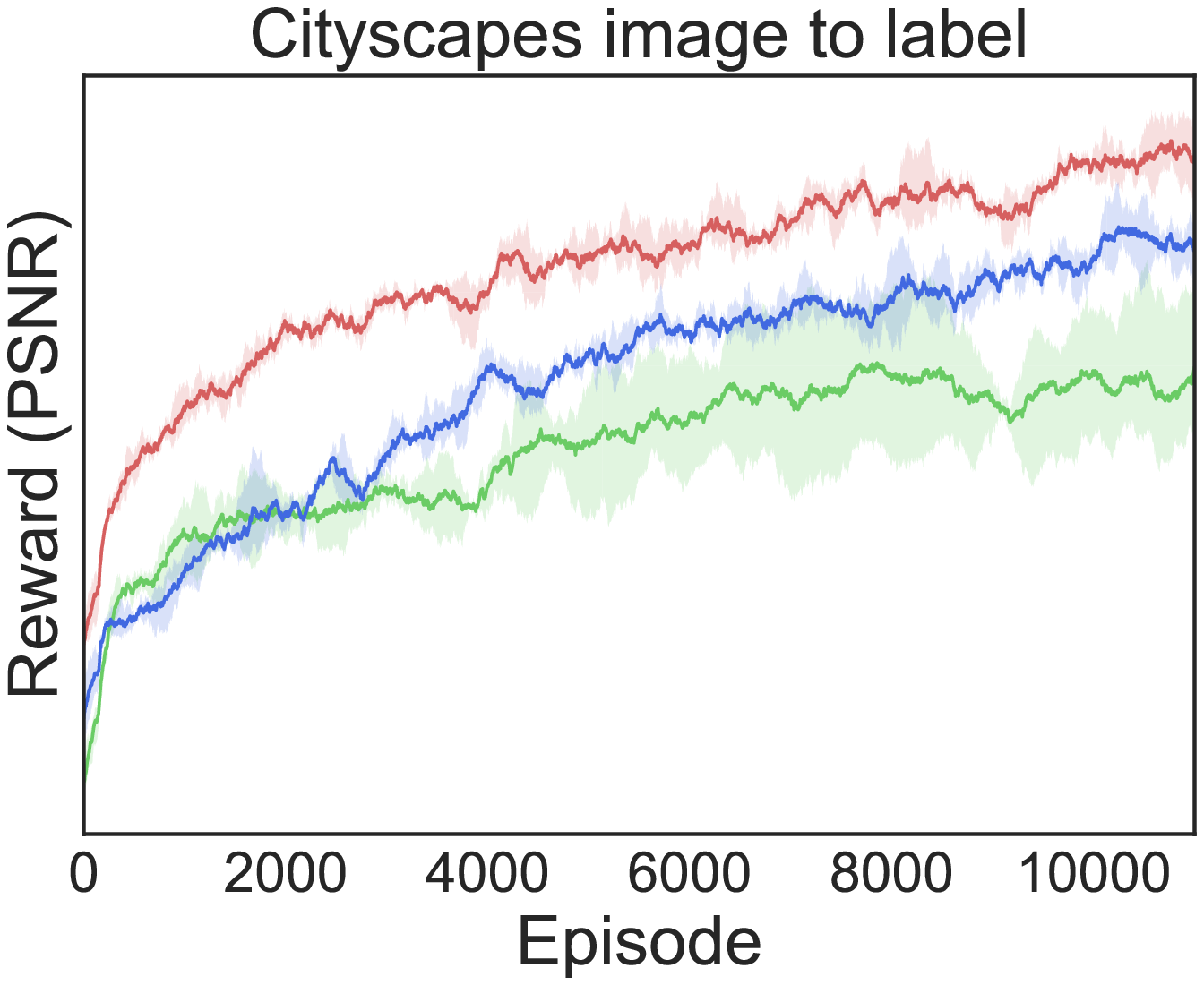}
 \includegraphics[width=0.24\textwidth]{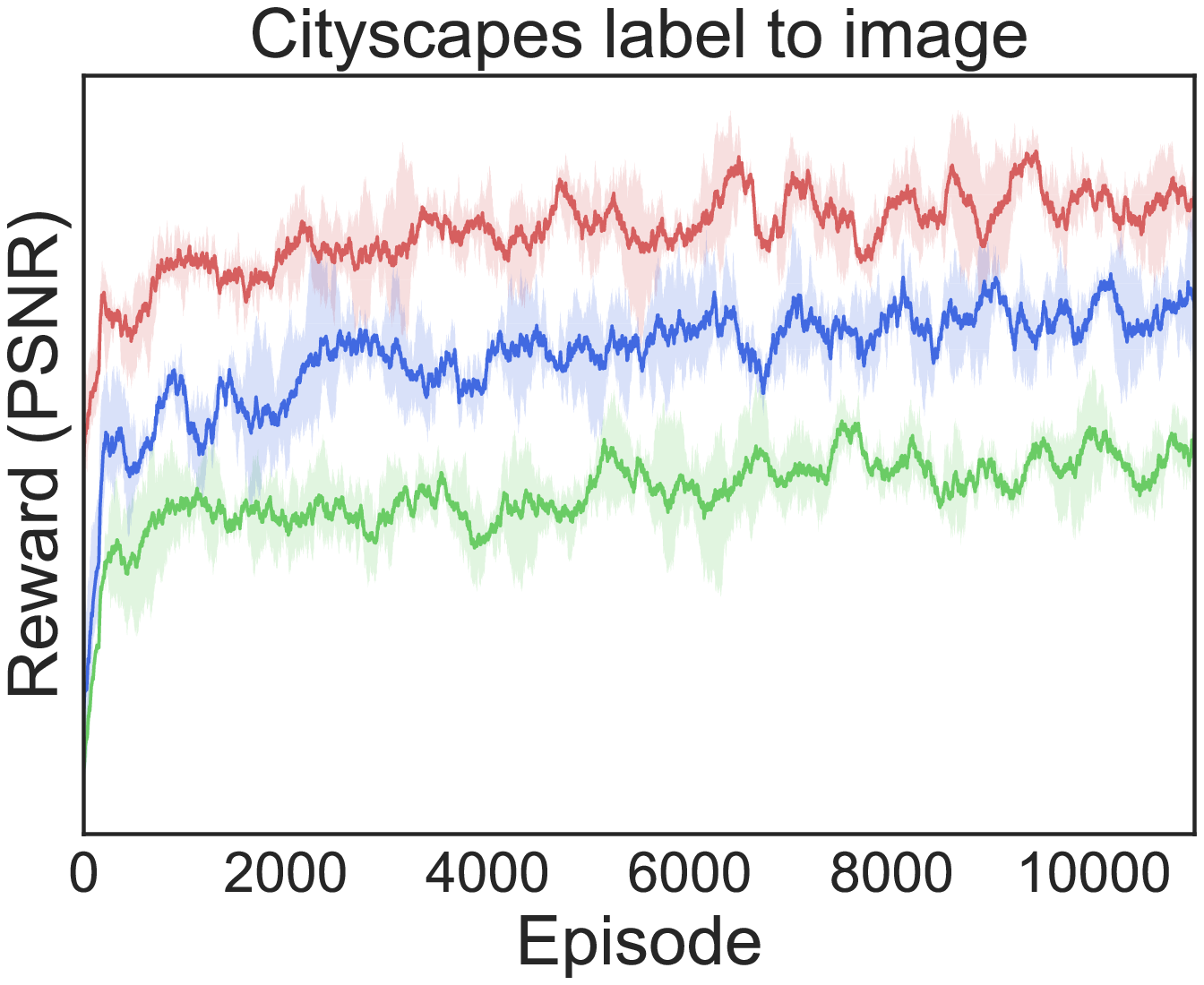}
 \includegraphics[width=0.24\textwidth]{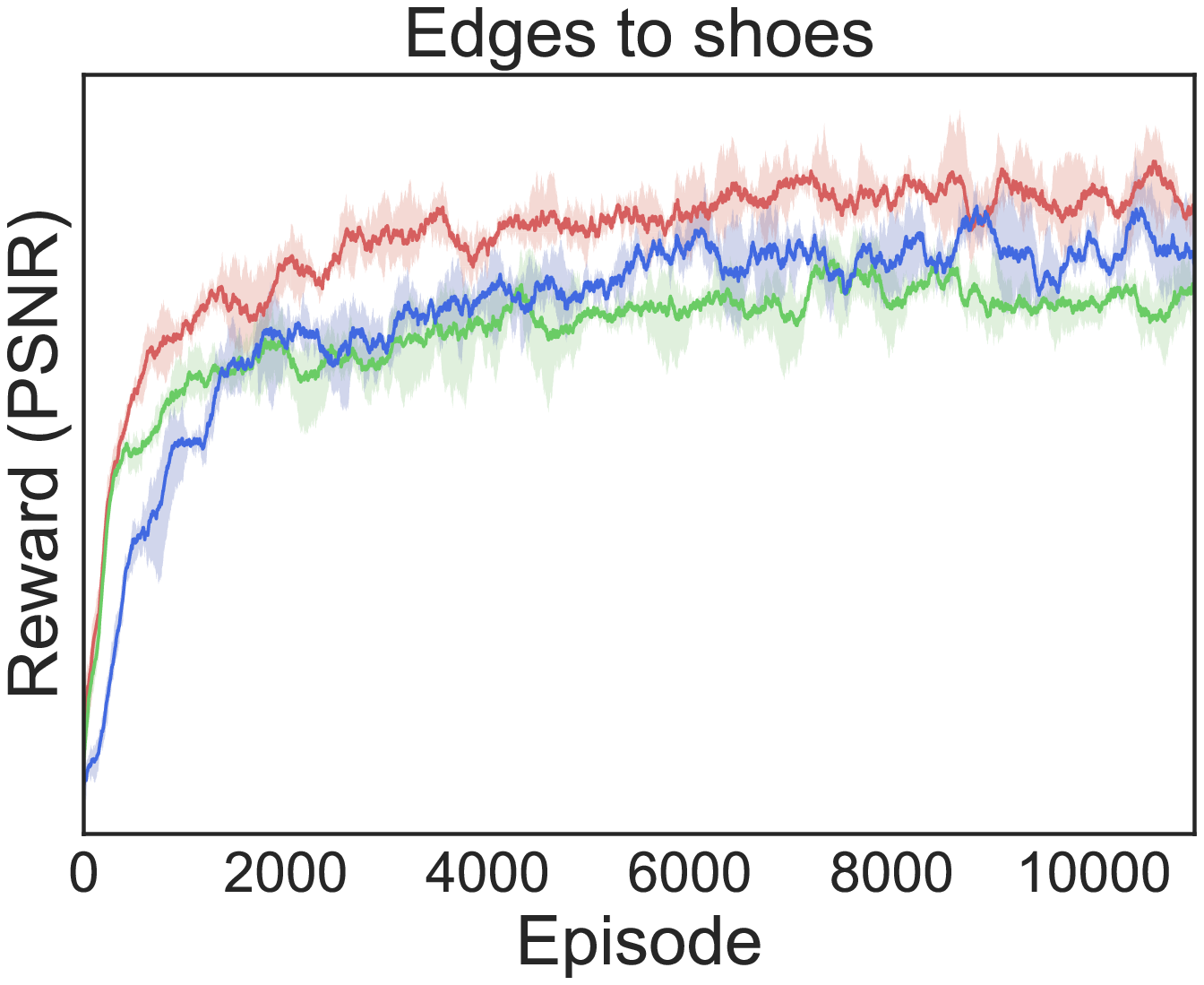}
}


\caption{Learning curves on different I2IT tasks. SAEC performs consistently better than other modified RL algorithms.}
\label{curves}
\end{figure*}

\noindent \textbf{Qualitative Evaluation}
The qualitative results of our SAEC with other I2IT methods on different tasks are shown in Figure \ref{l2i}. We observe that the pix2pix and PAN sometimes suffer from mode collapse and yield blurry outputs. The pix2pixHD is unstable in different datasets especially on Facades and Cityscapes. 
The DRPAN is more likely to produce blurred artifacts in several parts of the predicted image on Cityscapes. 
In contrast, the SAEC produces more stable and realistic results. Using stochastic policy and MERL helps to explore more possible solutions so as to seek out the best generation strategy by trial-and-error in the training steps, leading to a more robust agent for different datasets and tasks.

\noindent \textbf{Comparison with other RL Algorithms}
The key components of SAEC are substituted by other structures or other state-of-the-art RL algorithms to test their importance. 
We use DDPG and PPO to demonstrate the effectiveness of stochastic policy and maximum entropy RL, respectively.
The learning curves of different variants on four tasks are shown in Figure \ref{curves}.
By using the stochastic policy and the maximum entropy framework, the training is significantly improved.

\section{Conclusion}

In this paper, we present a new RL framework for high-dimensional continuous control and apply it to I2IT problems. This framework contains a {\em stochastic actor-executor critic} (SAEC) structure that handles high dimensional policy control, high-level state representation, and low-level image generation simultaneously. We train SAEC by jointly optimizing the RL and the DL objective, which provides ubiquitous and instantaneous supervision for learning I2IT. Our empirical results on several different vision applications demonstrate the effectiveness and robustness of the proposed SAEC. We will subsequently extend this work to more computer vision tasks involving with high-dimensional continuous spaces.

\section*{Acknowledgements}
This work was supported in part by the National Natural Science Foundation of China under Grant 61602065, Sichuan province Key Technology Research and Development project under Grant 2021YFG0038.


\end{document}